# NeuraHealth: An Automated Screening Pipeline to Detect Undiagnosed Cognitive Impairment in Electronic Health Records using Deep Learning and Natural Language Processing

Tanish Tyagi (1), Colin G. Magdamo (1), Ayush Noori (1), Zhaozhi Li (1), Xiao Liu (1), Mayuresh Deodhar (1), Zhuoqiao Hong (1), Wendong Ge (1), Elissa M. Ye (1), Yi-han Sheu (1), Haitham Alabsi (1), Laura Brenner (1), Gregory K. Robbins (1), Sahar Zafar (1), Nicole Benson (1), Lidia Moura (1), John Hsu (1), Alberto Serrano-Pozo (1), Dimitry Prokopenko (1 and 2), Rudolph E. Tanzi (1 and 2), Bradley T.Hyman (1), Deborah Blacker (1), Shibani S. Mukerji (1), M. Brandon Westover (1), Sudeshna Das (1)

(1) Massachusetts General Hospital, Boston, MA,

(2) McCance Center for Brain Health, Boston, MA

# Table of Contents

# Abstract

Dementia related cognitive impairment (CI) is a neurodegenerative disorder, affecting over 55 million people worldwide and growing rapidly at the rate of one new case every 3 seconds. 75% cases go undiagnosed globally with up to 90% in low-and-middle-income countries, leading to an estimated annual worldwide cost of USD 1.3 trillion, forecasted to reach 2.8 trillion by 2030. With no cure, a recurring failure of clinical trials, and a lack of early diagnosis, the mortality rate is 100%. Information in electronic health records (EHR) can provide vital clues for early detection of CI, but a manual review by experts is tedious and error prone. Several computational methods have been proposed, however, they lack an enhanced understanding of the linguistic context in complex language structures of EHR. Therefore, I propose a novel and more accurate framework, NeuraHealth, to identify patients who had no earlier diagnosis. In NeuraHealth, using patient EHR from Mass General Brigham BioBank, I fine-tuned a bi-directional attention-based deep learning natural language processing model to classify sequences. The sequence predictions were used to generate structured features as input for a patient level regularized logistic regression model. This two-step framework creates high dimensionality, outperforming all existing state-of-the-art computational methods as well as clinical methods. Further, I integrate the models into a real-world product, a web app, to create an automated EHR screening pipeline for scalable and high-speed discovery of undetected CI in EHR, making early diagnosis viable in medical facilities and in regions with scarce health services.

# 1 Introduction

## 1.1 Background

Dementia related cognitive impairment (CI) is a neurodegenerative disorder in which cells of the central nervous system progressively stop working and there is no cure. Common dementia related diseases include Alzheimer's, Parkinson's, Lewy Body, Huntington's, Vascular, Frontotemporal, and more. They affect over 55 million people worldwide, and are growing rapidly at the rate of one new case every 3 seconds, with an estimated annual worldwide cost reaching USD 1.3 trillion and forecasted to reach 2.8 trillion by 2030 (ADI, 2020; AA, 2016; WHO, 2012). Dementia has been recognized as a public health problem by the World Health Organization for over a decade and the rapidly aging global population is only compounding this problem (Prince, 2014; AA, 2020). In addition, over half of primary care physicians believe that they are not prepared for this growing problem (Amjad, 2018).

Despite high prevalence and key implications for patients and families, dementia is underdiagnosed by clinicians and underreported by patients and families (DementiaCareCentral, 2020). With 75% of dementia cases undiagnosed globally and up to 90% in low-and middle-income countries, most do not have access to treatment, care and organized support that getting an accurate formal diagnosis could provide (UPenn, 2021). Current methods take four to 52+ weeks for diagnosis causing additional burden on patients (SCIE, 2020). Therefore, when a diagnosis is made, the patient has often reached moderate dementia and irreversible damage has already been done to the brain (UWisconsinMadison, 2021). It robs a person of their identity, and eventually leads to death (100% fatality rate) (Khachaturian, 1985). With recurring failure of clinical trials, early detection of the first signs of CI is important for improving clinical outcomes and patient management. Tools that can efficiently and effectively analyze medical records for warning signs of dementia and recommend patients for follow up with a specialist can be critical to obtaining an early diagnosis for dementia. If these cases are diagnosed early, it enables the patient to utilize vital information and resources, including available drug and non-drug therapies for reversible symptoms, benefits to manage the devastating effect on their life, and helps them make important financial and legal decisions while the dementia is mild (UWisconsinMadison, 2021).

## 1.2 Research Problem

Current clinical methods to diagnose dementia utilize techniques such as computed tomography (CT), magnetic resonance imaging (MRI), and positron emission tomography (PET) scans (Khachaturian, 1985). These scans are very expensive and often not available to many people. CT scans cost $3,275, MRI scans cost $3,500, and PET scans cost upwards of $5,750 (AmericanHealthImaging, 2021; Deleon, 2022; Poslusny, 2018). Often, these tests need to be repeated, costing patients even more money and time, which is extremely of the essence in situations where it is likely the patient has dementia. This not only delays early detection and is cost prohibitive, but also, yields only 77% accuracy (Sabbagh, 2017).

Other methods such as cognitive and neuropsychological tests (NPT) evaluate a patient's thinking ability through testing memory, reasoning, problem-solving, language skills, visual and spatial skills, and other abilities related to mental functioning. However, NPT are time-consuming, require special training, and usually occur during a separate appointment with a neuropsychologist, in addition to an earlier visit to a regular neurology doctor (Stanford, 2021). NPT tests achieve only 80% accuracy (Jacova, 2007).

A review of patient's electronic health records (EHR) shows that clinicians may chart symptoms of cognitive issues in unstructured notes, but they may not make a formal diagnosis by entering it as a structured International Classification of Diseases (ICD) diagnosis code in a patient's EHR, refer to a specialist, or prescribe a medication (Boustani M C. C., 2005; Yarnall KS, 2003; AA, 2020; Bradford A, 2009; Boustani M P. A., 2006; Mowler NR, 2015).

With 89% adoption rate of EHR, its examination can provide vital clues for early detection of CI, which is essential to ensure patients get the right care and treatment to improve clinical outcomes (SCIE, 2020; Dugar, 2022). But a review of EHR by clinicians is manually intensive, time-consuming and error prone. These factors combined with a lack of time or expertise, patient resistance, and limited treatment options lead to dementia being severely underdiagnosed. Artificial intelligence (AI) based tool that can effectively and efficiently analyze medical records for warning signs of CI and recommend patients to follow up with a specialist is necessary in the fight against dementia.

### 1.3 Research Goals

This project develops a novel, state-of-the-art automated screening pipeline for scalable and high-speed discovery of undetected CI from EHR (unstructured clinician notes). Specifically, my main contributions are as follows:

- I developed a machine learning and an attention-based deep learning NLP model to understand the linguistic context from complex language structures to detect signs of CI in sequences (snippets) of EHR.
- I demonstrated that an improved understanding of the linguistic context and high dimensionality enhances the performance of CI classification, leading the deep learning NLP model to outperform the machine learning model for sequence classification.
- I developed a novel, two-step framework, NeuraHealth; first, predicting at the sequence level using deep learning NLP sequence classifier, and then, predicting at the patient level by applying logistic regression on sequence classifier to detect early signs of dementia in EHR. This framework significantly outperforms all state-of-the-art computational methods for patient level dementia detection as well as current clinical methods based on dementia-related ICD codes / medications in EHR.
- I integrated the NeuraHealth framework into a web application to create an automated screening pipeline that can be deployed at scale in medical facilities for high-speed and accurate discovery of undetected dementia by primary care physicians.

## 1.4 Related Works

Prior works have used NLP techniques to detect various diseases from EHR. (Rajkomar, 2018) used recurrent neural networks (long short-term memory (LSTM)) among others to predict inpatient mortality using EHR data from the University of California, San Francisco (UCSF) from 2012 to 2016, and University of Chicago Medicine (UCM) from 2009 to 2016. (Glicksberg et al., 2018) performed phenotyping for diseases such as attention deficit hyperactivity disorder (ADHD) by clustering on word2vec embeddings from EHR of the Mount Sinai Hospital (MSH) in New York City (Glicksberg, 2018). These studies have shown that the application of NLP techniques to EHR have improved disease detection, and that NLP techniques can be applied to dementia detection to achieve similar results. The current literature for dementia detection has utilized simple text-based analytics and machine learning methods, but they lack an enhanced understanding of the linguistic context in complex language structures of EHR, resulting in poor performance. My work uses novel state-of-the-art deep learning NLP techniques, that have achieved impressive results due to the use of word embeddings and attention-based bi-directional models (Vaswani, 2017; Mikolov, 2013; Pennington, 2014; Sarzynska-Wawer, 2021; Devlin, 2018), but have had limited applications in healthcare, and have not been hitherto applied to dementia detection. Another key difference in my work from the existing computational methods is the use of a two-step framework. I first classify sequences from EHR, then use the sequence predictions to generate structured features as input for a patient level regularized logistic regression.

## 2 Unstructured Data Preparation Pipeline

A database of ≈ 40,000 patients consisting of 10 million EHR from the Partners BioBank was filtered for age, Apolipoprotein E. (APOE) genotype data, and a match to dementia related keyword, resulting in extraction of 279,224 sequences from 16,428 unique patients. Select sequences were annotated by neurology physicians from Massachusetts General Hospital and always-patterns were developed to automate the annotation process for generating training data for the deep learning NLP model. The final dataset consisted of 8,656 labeled sequences from 2,487 patients and was used to develop a machine learning and attention-based deep learning NLP model.

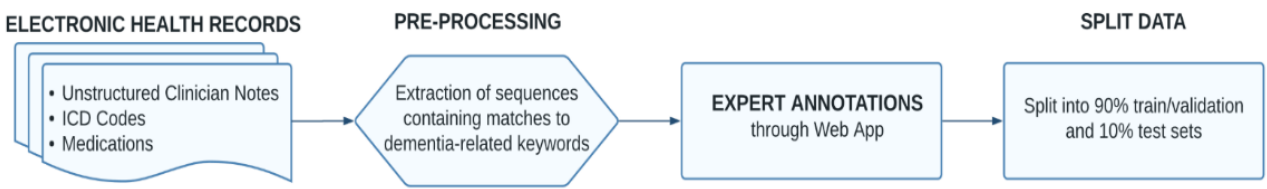

Figure 1: Unstructured Data Preparation Pipeline Overview Diagram

## 2.1 Dataset Description

The dataset originally consisted of ≈ 40,000 patients from the Partners BioBank, a Mass General Brigham (MGB) HealthCare (formerly Partner's Healthcare, comprising two major academic hospitals, community hospitals, and community health centers in the Boston area) initiative that houses genotype data for patients in the MGB Healthcare system.

The genotype of interest for this project was the APOE genotype, which is the biggest genetic risk factor for dementia (Mahley, 2000). The APOE genotype has 3 alleles: ε2, ε3, ε4. The ε2 allele is the rarest form of APOE and reduces the risk of developing dementia by up to 40%. ε3 is the most common allele and does not influence risk of dementia. The ε4 allele increases the risk for dementia and lowers the age of onset (Mahley, 2000). The APOE genotype data was used to ensure that the study consisted of diverse patients.

## 2.2 Dataset Filtering and Keyword Selection

The first step was to filter for patients who were older than 60 years of age (as of July 13, 2021) and who had an allele of the APOE genotype available in the BioBank, which resulted in an initial selection of ≈ 20K patients. I then developed a list of 18 dementia-related keywords (see Table 1) based on careful literature review of established methods for identifying patients with dementia using EHR (Gilmore-Bykovskyi, 2018). Expert neurologists at Massachusetts General Hospital (MGH) ensured that these keywords comprehensively capture evidence of CI, and that it would be exceedingly rare to describe CI (or the lack thereof) in EHR notes without using one of these keywords (Reuben D. B., 2017; Amra, 2017). Note that the presence of any of these keywords does not always indicate that the patient has CI.

| Number | Keyword | Match Count |
|---|---|---|
| 1 | Memory | 109218 |
| 2 | Cognition | 87655 |
| 3 | Dementia | 51034 |
| 4 | Cerebral | 45886 |
| 5 | Cerebrovascular | 36370 |
| 6 | Cerebellar | 26863 |
| 7 | Cognitive Impairment | 20267 |
| 8 | Alzheimer | 20581 |
| 9 | MOCA | 9767 |
| 10 | Neurocognitive | 7711 |
| 11 | MCI | 3889 |
| 12 | Amnesia | 3695 |
| 13 | AD | 2673 |
| 14 | Lewy | 2561 |
| 15 | MMSE | 2134 |
| 16 | LBD | 224 |
| 17 | Corticobasal | 147 |
| 18 | Picks | 41 |

Table 1: Keywords indicative of Cognitive Impairment

This list of keywords was used to further prune the dataset to only include patients who had at least one clinician note with a dementia-related keyword, which resulted in a final dataset consisting of 16,428 unique patients and 279,224 sequences. Table 2 shows the patient demographics for the cohort.

| Characteristic | (N = 16,428) |
|---|---|
| **Age (years) mean (SD)*** | 73.01 (7.96) |
| **Gender Male, *n* (%)** | 8740 (53.2) |
| **Race, *n* (%)** | |
| White | 14896 (90.7) |
| Other/Not Recorded | 608 (3.7) |
| Black | 570 (3.5) |
| Hispanic | 170 (1.0) |
| Asian | 168 (1.0) |
| Indigenous | 16 (0.01) |
| **APOE Genotype, *n* (%)** | |
| *APOE ε2* | 2028 (12.3) |
| *APOE ε3* | 10177 (62.0) |
| *APOE ε4* | 4223 (25.7) |
| **Average Speciality Visits (SD)** | 1.67 (4.6) |
| **Average PCP Encounters (SD)** | 5.25 (5.63) |

Table 2: Patient Demographics of the Dataset, *SD: Standard Deviation

## 2.3 Sequence Construction and Extraction from Dataset

For each patient in the dataset, I extracted unstructured clinician notes, identified matches to 18 dementia-related keywords (Table 1), including those related to memory, cognition, neuropsychological tests, and dementia diagnoses. I constructed sequences from the note text spanning each of these matches (of length 800 characters). The below preprocessing steps were followed to construct sequences that could be easily interpreted by humans and the models:

a. Removed all empty lines and multiple white spaces.
b. Computed context windows of 100 characters before start of keyword match and 100 characters after.
c. For notes that had multiple keyword matches, the context windows were merged.
d. Constructed sequences by extracting note text from computed context windows.
e. Tokenized extracted sequences into BERT tokens (where 1 token = 1 word) and extended context windows for all sequences that were less than 512 tokens.

f. Cleaned up spaces and other special characters through regular expression substitutions to make the sequence more readable for human annotators.

The final cohort of 16,428 patients had 279,224 total sequences constructed with dementia-related keywords. Table 3 shows summary statistics of the sequences.

| Characteristic | (N = 279,224) |
|---|---|
| Average Sequence Length (SD) | 910 (485) |
| Average Keyword Count (SD) | 1.97 (1.62) |
| % Sequences with 1 Keyword Match | 54.5 |
| % Sequences with 2 Keyword Matches | 24.2 |
| % Sequences with 3 Keyword Matches | 9.30 |
| % Sequences with 4+ Keyword Matches | 12.0 |

Table 3: Summary statistics of Sequence Cohort

## 2.4 Sequence Annotation

A subset of 5,000 sequences from 5,000 unique patients was generated for annotation (labeling) such that the relative frequency of the keywords in the subset is the same as that in the original set of 279,224 sequences. A subset of the selected sequences was annotated for indication of CI. In this context, CI was defined as evidence of either Mild CI (MCI), where one cognitive domain is involved, or dementia, where more than one cognitive domain is involved, and activities of daily living are affected. Concern from the family of the patient or the patient was not considered as CI.

**Classification Task:** Each sequence was labeled with one of three classes:
1. Positive, i.e., patient has CI
2. Negative, i.e., patient does not have CI
3. Neither, i.e., sequence does not contain information pertinent to a patient's cognitive status

Sequences were annotated using two approaches: One, manually by neurology physicians from Massachusetts General Hospital (MGH). Two, by utilizing an always-pattern scheme to automate the annotation process. An always pattern was defined as a phrase or regex expression that in any context indicates the phrase will be labeled with a particular class (i.e., positive, negative, or neither). Once an always pattern is defined, all other sequences that match the pattern are automatically labeled with that always pattern's class. Figure 2 contains examples of sequences and always patterns for all three classes.

| Positive Sequences | Positive Always Patterns |
|---|---|
| 1. Patient MOCA is 22/30. | 1. (?i)\bMOCA\s*([0-9]\|[12][0-5])\s*/\s*30 |
| 2. Patient with past medical history of dementia. | 2. (?i)\bpast\s*medical\s*history\s*[^.]*(dementia) |

| Negative Sequences | Negative Always Patterns |
|---|---|
| 1. Patient memory is intact. | 1. (?i)Memory.*intact |
| 2. No memory concerns. | 2. (?i)No\s*memory\s*concerns |

| Neither Sequences | Neither Always Patterns |
|---|---|
| 1. History: Father has Alzheimer's Disease | 1. (?i)Father.*Alzheimer's\s*disease |
| 2. Patient attends anticoagulation therapy daily. | 2. (?i)anticoagulation |

Figure 2: Example Sequence and Always Patterns

Both, manual and automated annotations were carried out by experts using a web-based annotation tool. The tool was constructed using Python-based open-source Django web development framework with an SQLite database. Data Models were established for the selected sequences, clinician notes, user account creation and authentication, and sequence assignment to individual or multiple annotator accounts. User interface (UI) screens were created to present the data to the annotators and for them to execute the annotation process.

The manual annotation by Neurologists and the automated annotation, both executed via the web tool, results in a final dataset of 8,656 annotated sequences from 2,487 unique patients was split between training and validation set (90%), and holdout test set (10%). The training, validation, and holdout test sets were stratified across label and proportion of sequences annotated manually and through always patterns. No patients were featured in multiple sets.

## 3 Sequence Classification Methods (Step 1)

Two NLP models were developed for the sequence classification task and compared to each other: a baseline TF-IDF (term frequency-inverse document frequency) machine learning model and a fine-tuned attention-based ClinicalBERT deep learning model.

## 3.1 Logistic Regression with TF-IDF Vectors

TF-IDF vectorization was performed on the annotated 8,656 sequences and feature selection was based on a term's Pearson correlation coefficient (PCC) with the CI label (Ramos, 2003; Benesty, 2009). L1 Regularized Logistic Regression was applied with the annotated CI labels (Tibshirani, 1996). A 10-Fold Cross Validation was used to determine optimal hyperparameters (lambda value and correlation coefficient threshold) to select features (Refaeilzadeh, 2009). Figure 3 depicts the procedure for the machine learning model.

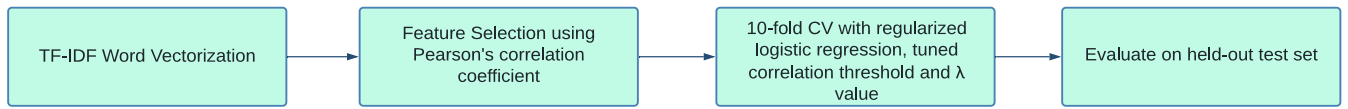

Figure 3: Machine Learning Model Overview Diagram

First, annotated sequences are converted into TF-IDF vectors. This is done through TF-IDF vectorization, which converts the text into a vector by counting the occurrence of words in a document. TF-IDF vectors take this a step further as they contain insights about the less relevant and more relevant words in a document, which is of great significance.

### 3.1.1 TF-IDF Computation

For a particular word, the TF-IDF value is the product of the term frequency ($TF$) and inverse document frequency ($IDF$). TF is of the frequency of a word ($w$) in a document ($d$).
$TF(w, d) = \frac{frequency\ of\ w\ in\ d}{total\ \#\ of\ words\ in\ d}$.

IDF measures the importance of each word, and provides a weightage based on the frequency of a particular word ($w$) in the corpus (collection of documents) ($c$). $IDF(w, c) = \ln\left(\frac{total\ \#\ of\ documents\ in\ c}{\#\ of\ documents\ containing\ w}\right)$. Therefore, $TF - IDF(w, d, c) = TF(w, d) * IDF(w, c)$. TF-IDF creates a vector for each document in the corpus that has dimensions $1 * length\ of\ vocabulary\ (total\ words\ in\ corpus)$.

A limitation of TF-IDF is that it can be computationally expensive for large vocabularies. To combat this, I eliminated word features that were deemed to have little correlation to the cognitive impairment label using the PCC. PCC is the measure of correlation between two sets of data and ranges from 0 - 1. It is defined as the ratio between the covariance of two variables and the product of their standard deviations, and is defined below:

$$For\ paired\ data\ \{(x_1, y_1), (x_2, y_2), \ldots, (x_n, y_n)\}, PCC(x, y) = \frac{\sum_{i=1}^{n}(x_i - \bar{x})(y_i - \bar{y})}{\sqrt{\sum_{i=1}^{n}(x_i - \bar{x})^2} \sqrt{\sum_{i=1}^{n}(y_i - \bar{y})^2}}.$$

Once the PCC was computed for each of the word features, a L1 Regularized Logistic Regression model was regressed on the TF-IDF vectors. Logistic Regression is a regression technique is an adaption of linear regression to create a classification model. It is defined as: $h_\theta(x) = \frac{1}{1 + e^{-(\theta_0 + \theta_1 X_1 + \theta_2 X_2 + \theta_3 X_3 + \ldots + \theta_n X_n)}}$ with a cost function defined as

$$J(\theta) = -\frac{1}{m}\left[\sum_{i=1}^{m} y^i \log\left(h_\theta\left(x^i + (1 - y^i)\right)\right) \log\left(1 - h_\theta(x^i)\right)\right]$$

plus a $\frac{\lambda}{2m} \sum_{j=1}^{n} \theta_j^2$ regularization parameter to reduce overfitting. 10-fold cross validation was used to tune for a PCC threshold that removed the optimal amount of word features to maximize the performance metrics of a L1 Regularized logistic regression model that was regressed on the TF-IDF vectors where no element had a PCC score less than the arbitrary threshold.

In the 10-fold cross validation loop, the training data (7,487 annotated sequences) was split into 10 subsets. A holdout procedure then commenced for 10 iterations, where for each iteration, one of the subsets was chosen as a validation set while the other 9 formed the training set. The validation set was used to tune the hyperparameters for the L1 Regularized logistic regression model, specifically the λ value, which controls the impact of the regularization parameter on the cost function, and PCC threshold. The model trained on the 9 subsets and evaluated itself on the validation set, adjusting the hyperparameters.

## 3.2 Attention-based Deep Learning ClinicalBERT Model

I utilized a pre-trained language model called ClinicalBERT, which was trained on the MIMIC II database containing EHR records from ICU patients (Alsentzer, 2019; Saeed, 2011). The model was programmed using the implementation available in the Huggingface Transformers and Simpletransformers packages (Wolf, 2019; Rajapakse, 2019). After text preprocessing, input texts were tokenized with the default tokenizer and converted to embeddings. The model was initialized with pre-trained parameters and later fine-tuned on my labeled training set. Adam

Optimizer and Optuna were used to perform a 20-trial study and tune the learning rate, Adam Epilson, and the number of train epochs on the held-out validation set (Kingma, 2014; Akiba, 2019). An early stopping rule was used to prevent overfitting by ensuring that training stopped if the loss did not change substantially over 3 epochs. Figure 4 depicts the procedure for the deep learning model.

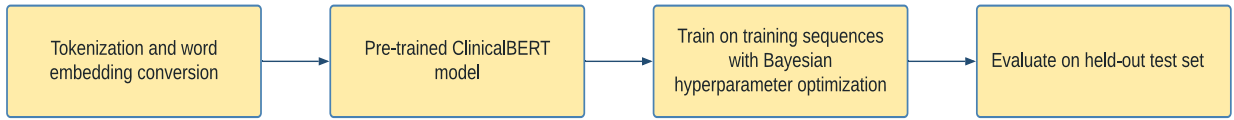

Figure 4: Deep Learning Model Overview Diagram

### 3.2.1 ClinicalBERT Model Architecture

Figure 5 shows the architecture for the ClinicalBERT model. ClinicalBERT has a transformer architecture, which enables models to process text in a bidirectional manner, from start to finish and from finish to start (Vaswani, 2017). This design overcomes the limits of previous state-of-the-art models such as Long short-term memory (LSTM) models, which could only process text from start to finish. The scaled dot-product attention and multi-head attention layers capture the relationships between each word in a sequence with every other word, which allows ClinicalBERT to achieve higher performance levels than the TF-IDF approach.

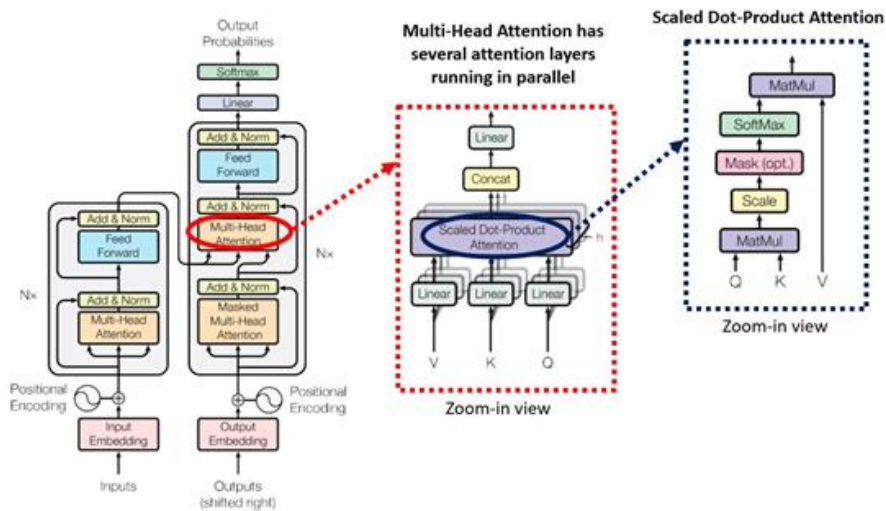

Figure 5: ClincalBERT's Transformer Architecture, Source: (Vaswani, 2017)

### 3.2.2 ClinicalBERT Model Computation

The input into the scaled dot-production attention layer consists of queries ($Q$) and keys ($K$) of dimension $d_k$, and values ($V$) of dimension $d_v$. The dot products of the query with all keys are computed, divided by $\sqrt{d_k}$, and followed by the application of the softmax ($\sigma$) function to obtain the weights on the values. $\sigma(\vec{z})_i = \frac{e^{z_i}}{\sum_{j=1}^{K} e^{z_j}}, attention(Q, K, Z) = \sigma(\frac{QK^T}{\sqrt{d_k}})V$.

The Multi-head Attention layer is a module for attention mechanisms which runs through an attention mechanism several times in parallel. The independent attention outputs are then concatenated and linearly transformed into the expected dimension. Multiple attention heads allow for attending to parts of the sequence differently.

$MultiHead(Q, K, V) = [head_0, head_1, head_2, \ldots, head_h]W_0$ where $head_i = attention(QW_i^Q, KW_i^K, VW_i^V), W = learnable\ parameter\ matrices$. For more details regarding the transformer architecture, see (Vaswani, 2017).

ClinicalBERT was initialized from the transformer model and trained on the MIMIC II database containing EHR records from ICU patients. This training allowed the model to develop an understanding on clinical terminology. Since the attention mechanism in the Transformer allows ClinicalBERT to model any downstream task, I fine-tuned it on my training set so that it could develop an understanding of terminology relevant to Cognitive Impairment.

The held-out validation set was used to tune the hyperparameters learning rate, Adam Epilson, and the number of train epochs. To tune these hyperparameters, the Bayesian hyperparameter optimization library Optuna was used. Optuna employs a pruning strategy that constantly checks for algorithm performance during training and terminates a trial if a combination of hyperparameters does not yield good results, and a sampling algorithm for selecting the best hyperparameter combination, concentrating on hyperparameters which yield good results and ignoring those that do not. I created a 20 trial Optuna study designed to maximize accuracy with Tree-structured Parzen Estimator (TPE) sampling algorithm (Bergstra, 2011). The learning rate and Adam Epilson were tuned from ranges of [1e$^{-8}$, 1e$^{-4}$], and number of training epochs was tuned between 1 and 3.

I trained the ClinicalBERT model on a Linux cloud cluster using two 16GB NVIDIA Graphic Processing Units (GPUs) over a period of 25 hours.

# 4 Performance of Sequence Classification Models

I evaluated each of the two models, TF-IDF and ClinicalBERT, based on sequence level class assignments. Model performance on the held-out test set is shown in Table 4. To compute each metric, I used the threshold that maximized accuracy.

| Model | Accuracy | AUC | Sensitivity | Specificity | Weighted F1 |
| --- | --- | --- | --- | --- | --- |
| TF-IDF | 0.85 | 0.94 | 0.83 | 0.92 | 0.84 |
| ClinicalBERT (fine-tuned for dementia) | 0.93 | 0.98 | 0.91 | 0.96 | 0.93 |

Table 4: Performance of the two models I trained for Sequence Classification

## 4.1 TF-IDF Performance

The TF-IDF model achieved an AUC of 0.94, accuracy of 0.85, sensitivity of 0.83, specificity of 0.92, and weighted F1 of 0.84. Hyperparameters were selected by finding the combination of the λ value and PCC threshold that maximized the average accuracy over the 10 CV folds. The optimal λ value and PCC threshold were 10 and 0.01, respectively. Word features related to memory and CI had the highest coefficients in the model. The 20 words with the highest correlation coefficients using TF-IDF word vectorization are shown in Table 5.

| Number | Word | Correlation | Number | Word | Correlation |
| --- | --- | --- | --- | --- | --- |
| 1 | Intact | 0.5573 | 11 | Homicidal | 0.3610 |
| 2 | Oriented | 0.4233 | 12 | Observation | 0.3602 |
| 3 | Concentration | 0.4157 | 13 | Knowledge | 0.3598 |
| 4 | Orientation | 0.4029 | 14 | Insight | 0.3561 |
| 5 | Perceptions | 0.3959 | 15 | Associations | 0.3538 |
| 6 | Sensorium | 0.3954 | 16 | Abstract | 0.3524 |
| 7 | Judgement | 0.3851 | 17 | Suicidal | 0.3514 |
| 8 | Fund | 0.3733 | 18 | Attention | 0.3433 |
| 9 | Experiences | 0.3693 | 19 | Content | 0.3396 |
| 10 | Ideation | 0.3612 | 20 | Thought | 0.3385 |

Table 5: Top 20 TF-IDF Word Features and their PCC

The precision matrix and ROC curve for TF-IDF can be found in Figures 6 and 7 respectively.

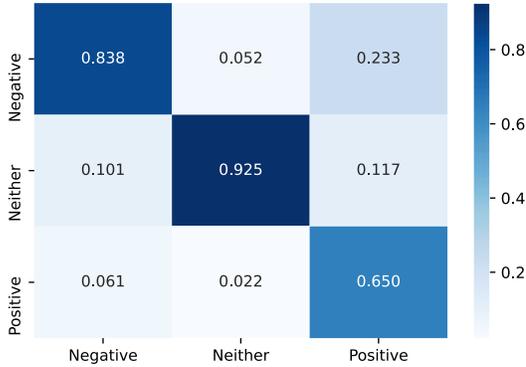
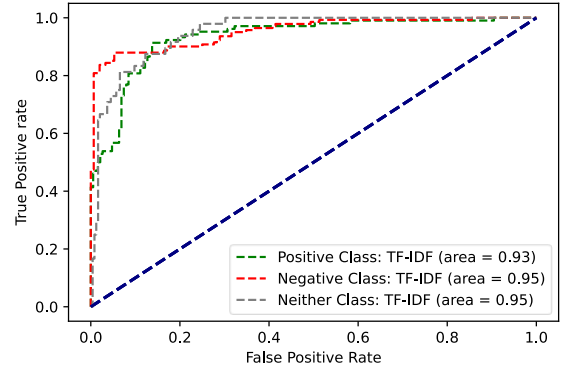

Figure 6: TF-IDF Precision Matrix            Figure 7: TF-IDF ROC Curve

While TF-IDF was able to identify the presence of a keyword or always pattern in a sequence, it was unable to the leverage the context around each keyword match. The context of the keywords and the agents within the sentence often contained useful information regarding a patient's cognitive status. For example, the sequence "Patient is caregiver for wife who has dementia" has the keyword dementia, but it does not pertain to the patient's cognitive diagnosis but instead their wife's. This led the baseline TF-IDF model to incorrectly predict sequences as evidence of cognitive impairment, resulting in a large count of false positives, as shown by the precision matrices in Figure 6.

## 4.2 ClinicalBERT Performance

ClinicalBERT, with its more complex architecture as discussed in Section 3.2, was able to leverage the context of the keyword matches within the sequences and overcome the issues faced with TF-IDF performance. This was evident in the results, as the fine-tuned ClinicalBERT model achieved an AUC of 0.98 and substantially improved accuracy to 0.93 as well as specificity of 0.96, sensitivity of 0.91, and weighted F1 of 0.93. The precision matrix and ROC curve for ClinicalBERT can be found in Figures 8 and 9, respectively.

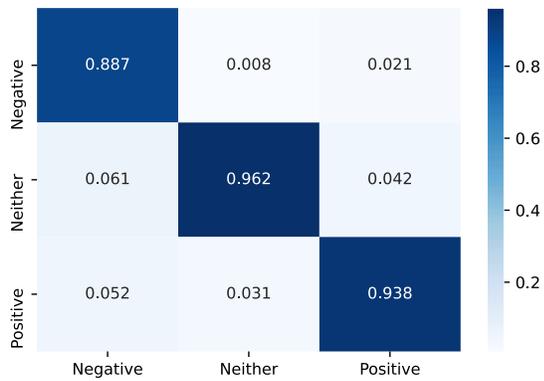
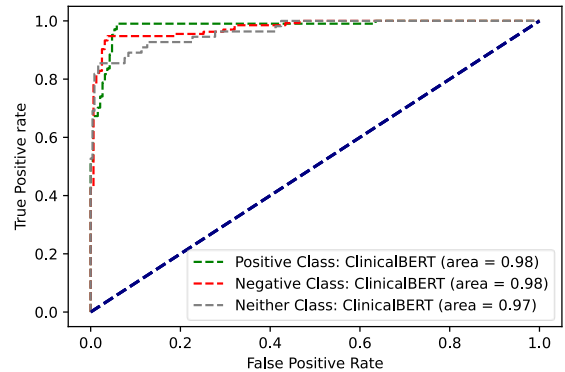

Figure 8: ClinicalBERT Precision Matrix

Figure 9: ClinicalBERT ROC Curve

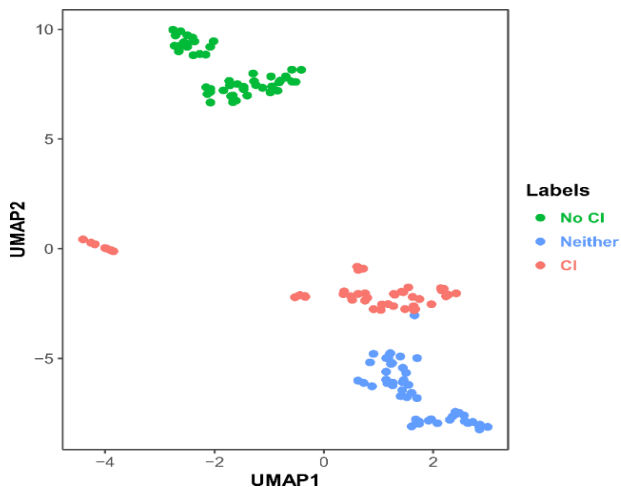

Additionally, when using a small dataset of manually annotated sequences (N = 150) which did not match an always pattern, ClinicalBERT was able to accurately discriminate between all three classes (see Figure 10).

This experiment further demonstrates the high accuracy of my fine-tuned deep learning model for sequence classification.

Figure 10: UMAP Clustering of ClinicalBERT Embeddings

Examples of sequences with the keyword "memory" used in different contexts:

**(1) No:** "fund of knowledge and memory were normal"

**(2) Neither:** "mother had memory problems in her 70s"

**(3) Yes:** "increased short-term memory loss and confusion"

# 5   Patient Classification Methods (Step 2)

To make this project fully applicable in a clinical setting, it would need to return an overall prediction regarding whether the patient had CI or not (not just at the Sequence level but also at the Patient level). However, annotators had only annotated whether a particular sequence showed signs of a patient having CI in the BioBank dataset. To get ground truth labels at the patient level, I utilized another in-house dataset curated in (Hong, 2020), where each patient's EHR record between 01/01/2018 – 12/31/2018 was reviewed by an expert clinician (neurologist, psychiatrist, or geriatric psychiatrist) to label patients with presence or absence of cognitive impairment. This dataset is annotated at the patient level versus the dataset I used earlier in Sections 3 is annotated at the sequence level. After running this dataset through the Data Preparation Pipeline described in Section 2, a gold-standard patient level dataset from 921 unique patients containing 46,650 sequences was created. These patients / sequences are completely new dataset and were not part of the sequence level dataset that ClinicalBERT was trained and evaluated on in Section 3.2.

ClinicalBERT was then applied to this dataset to generate the sequence level predictions. Using these predictions, four structured features were generated per patient: percent sequences predicted positive, percent sequences predicted negative, percent sequences predicted neither, and total sequence count.

I then filtered for all patients who had greater than 10 annotated sequences to create a final patient level dataset consisting of 691 patients. After applying feature standardization, defined as $X' = \frac{X-\mu}{\sigma}$, data was split with train (90%) and holdout test (10%) sets. Validation datasets were split from the train set using techniques described in the Section 2.4. A L1 Regularized logistic regression model was regressed on these features with the patient level CI label as the outcome. To tune hyperparameters, specifically the λ value, a 10-fold cross validation loop was used.

## 5.1  Patient Classification Model Performance

When evaluated on the hold-out test set, the patient level model achieved an accuracy of 0.88, AUC of 0.92, Sensitivity of 0.91, Specificity of 0.90, and F1 of 0.89. The precision matrix and ROC curve for the patient level model can be found in Figures 11 and 12, respectively.

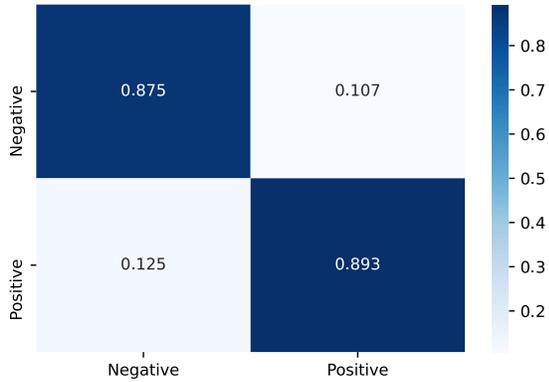
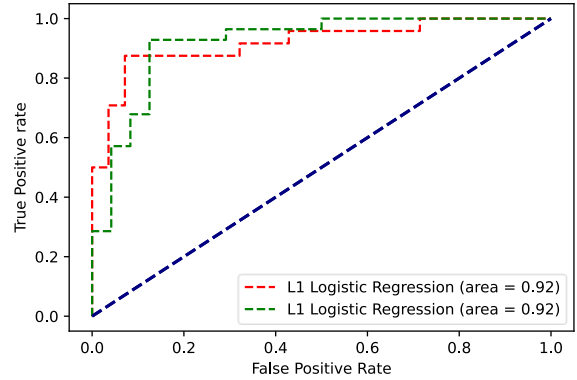

Figure 11: Patient Level Model Precision Matrix

Figure 12: Patient Level Model ROC Curve

## 5.2 Comparison with Current Computational Methods

I compared the performance of my method with all existing methods as evaluated upon accuracy, AUC, sensitivity, and specificity. The methods fall into three main approaches currently being used for dementia detection using EHR data: text-based analytics, machine learning, and deep learning

The results are in Table 9. My method significantly outperforms all other methods, with respect to accuracy, AUC, sensitivity, and specificity improvements of 15%, 7%, 8%, and 23% respectively. This can be attributed to the state-of-the-art ClinicalBERT model architecture that builds an understanding of the linguistic context in complex language structures of EHR data, as well as my two-step framework of predicting on the sequence level before aggregating on the patient level.

| Methods | Accuracy | AUC | Sensitivity | Specificity |
|---|---|---|---|---|
| **Text-Based Analytics (Reuben D. B., 2017)** | — | — | — | 65% |
| **Machine Learning Methods** | | | | |
| TF-IDF (Searle, 2020) | 73% | | 83% | |
| Support Vector Machines (SVM) (Battista, 2017) | | | | |
|     SVM-Linear | 65% | — | 59% | 67% |
|     SVM-Quadratic | 65% | — | 59% | 67% |
|     SVM-Gaussian | 64% | — | 61% | 67% |
| Logistic Regression (So, 2017) | — | — | 78% | — |
| Naïve Bayes (Garrard, 2014) | 68% | — | — | — |
| LightGBM **(Hane, 2020)** | — | 85% | — | — |
| Decision Trees **(Luz, 2020)** | 63% | — | — | — |
| Linear Discriminant Analysis (LDA) **(Luz, 2020)** | 61% | — | 78% | — |
| **Deep Learning Methods** | | | | |
| Convolutional Neural Network Long Short-Term Memory Network (CNN-LSTM) **(Karlekar S., 2018)** | 69% | — | 69% | — |
| Recurrent Neural Network (RNN) **(Nori, 2020)** | — | 81% | — | — |
| Feed Forward Network **(Nori, 2020)** | — | 81% | — | — |
| **NeuraHealth: ClinicalBERT+Logistic Regression (Mine)** | **88%** | **92%** | **91%** | **90%** |

Table 9: Comparison of NeuraHealth showing that it significantly outperforms other state-of-the-art Computational Method for detection of CI in EHR data

## 5.3 Comparison with Current Clinical Methods

To further see the utility of my method over current clinical methods, I compared the patient level predictions of the 52 patients in the test-set to the output of current clinical methods for dementia diagnosis for these patients. While my method accurately predicts the dementia status for 88% of the patients, current clinical methods correctly predict only 77% patients (Sabbagh, 2017), and neuropsychological tests (NPT), achieve only 80% accuracy for early detection (Jacova, 2007). The comparison is in Table 10.

| **Current Clinical Methods** | **Accuracy** |
|---|---|
| CT Scans **(Sabbagh, 2017)** – Very late diagnosis | 77% |
| NPT Tests **(Jacova, 2007)** –Late diagnosis | 80% |
| **NeuraHealth: ClinicalBERT+Logistic Regression – Early diagnosis (Mine)** | **88%** |

Table 10: Comparison of NeuraHealth (my method) showing that it also outperforms current Clinical methods for detection of CI

That means my method has the potential of accurately classifying up to 11% more patients than current clinical methods. When the fact that dementia is a rapidly growing problem that will affect nearly 80 million people in 2030 (WHO, 2021) and that 75% of dementia cases currently go undiagnosed is considered, my method can have an enormous impact if deployed, which I discuss in the following section.

These results are a major improvement over current clinical methods, which are only able to achieve 77% accuracy (Sabbagh, 2017). As shown, the patient level model was able to identify a significant proportion of patients that went undetected by current clinical methods, highlighting the utility of such a tool in a clinical setting.

# 6 System Integration with Web Application for Scaling Early Detection

I deployed the NeuraHealth Patient Classifier as a web application designed for real-world use by both doctors and patients. I integrated my ClinicalBERT model and Logistic Regression Patient Level classifier developed in python with an Apache web server using the Django framework. I also add a frontend interface for the required inputs for using the model. This allows the seamless use of the model by any user without working directly with the underlying python implementation.

The Django web framework was used to develop the backend and frontend components of the application. To detect CI using the web app, the user uploads a file containing EHR notes. The application first runs the sequence extraction pipeline to extract sequences of interest, which are then feed to the fine-tuned ClinicalBERT model to generate sequence level predictions. These predictions are converted to features (as described in Section 5) and fed into the patient level classification model for overall prediction of CI.

Following this, the user receives a comprehensive report detailing their cognitive impairment status and a recommendation from the model on future next steps. The report includes the overall patient level CI probability, the number of sequences with a high probability of CI (>50%), a scatterplot showing the correlation between their sequences and probability of CI, and model predictions / three-class probabilities for each sequence, which is displayed with its keywords highlighted for convenience (Figure 13 and 14).

NeuraHealth web application processes the patient's EHR notes, runs the models; both Sequence Classifier and Patient Classifier, and generates summary and detailed results showing the patient's probability of CI in less than 5 seconds. Figure 13 shows the Web app powered by NeuraHealth. It demonstrates that a healthcare facility can screen all their patients for CI every month, automatically, and receive a recommended list of patients that must visit a specialist. This outperforms current clinical methods by 11%.

NeuraHealth is integrated into a web app to create an automated screening pipeline for scalable and high-speed discovery of undetected CI in EHR. Plus, my diagnostic method is cheaper, faster, and more accessible making early diagnosis viable in medical facilities and in regions with scarce health services.

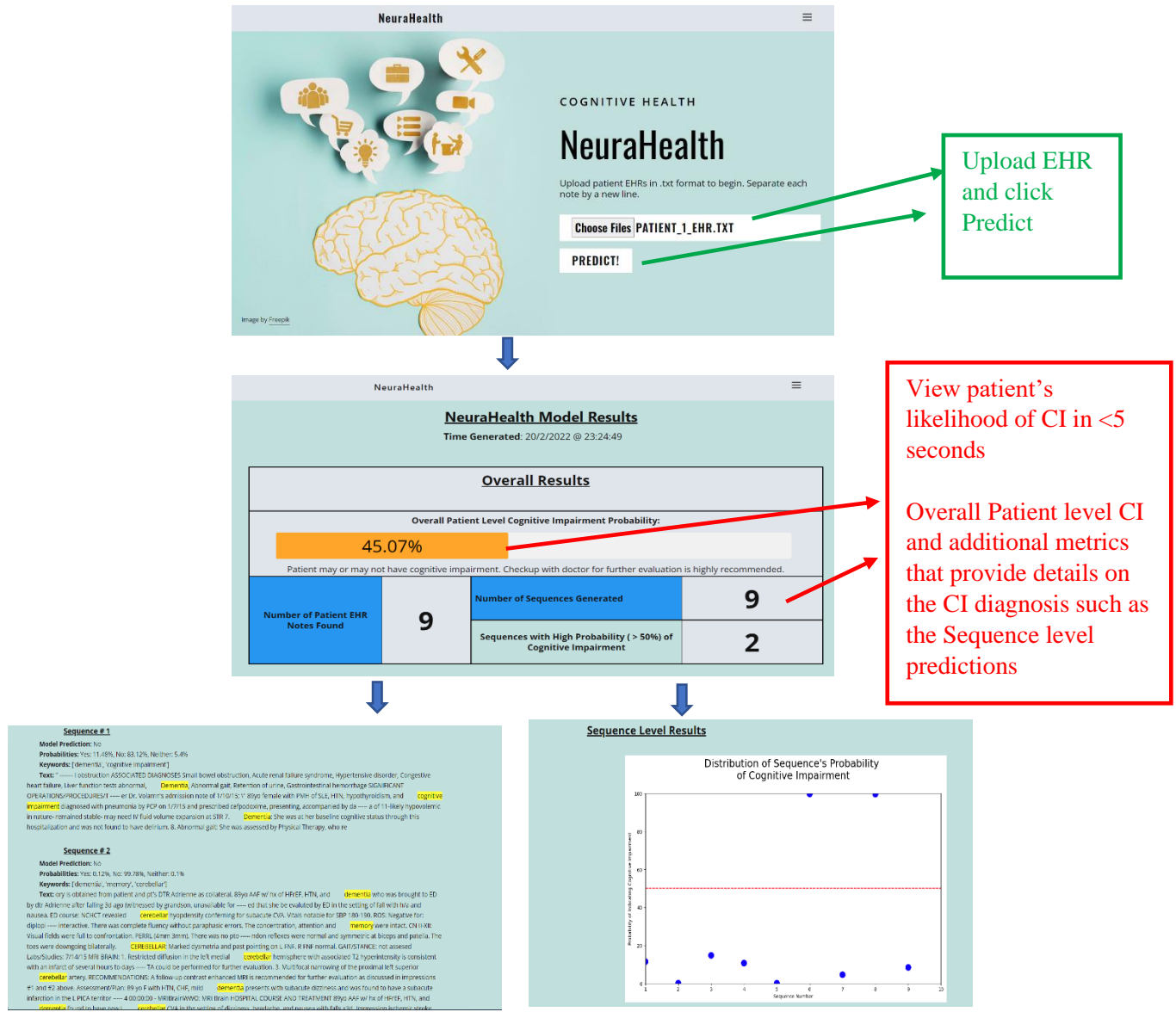

Figure 13 shows the web app powered by NeuraHealth

# 7   Conclusion, Application, and Future Work

I applied NLP algorithms to identify patients with cognitive impairment in EHR and compared a baseline TF-IDF model with an attention based deep learning model, ClinicalBERT, for sequence level class assignment predictions. The deep learning model fine-tuned to understand dementia was able to leverage the linguistic context in complex language structures of EHR and significantly outperformed on all fronts. My work illustrates the need for more complex, expressive language models for the nuanced task of detecting dementia related cognitive impairment in electronic health records.

NeuraHealth, a novel two-step framework I developed to predict a patient's overall cognitive status, uses the deep learning model to first classify sequences from EHR and then uses sequence predictions as an input to a patient level regularized logistic regression model, creating high dimensionality. I demonstrate that NeuraHealth outperforms all existing state-of-the-art computational methods.

By integrating NeuraHealth in a web application, I demonstrate that it can be deployed at scale in medical facilities for primary care physicians to manage clinical outcomes. This work can help address the underdiagnosis of dementia and alert primary care physicians to do a formal cognitive evaluation or refer to specialists. Such a tool can be used to facilitate real-world research to generate cohorts for dementia studies to identify risk and protective factors of dementia as well as recruit patients into observational studies or clinical trials.

In summary, I created an automated screening pipeline to detect undiagnosed dementia in electronic health records and the methods used in NeuraHealth are faster, cheaper, and more accurate than all current state-of-the-art computational and clinical methods. My work was able to outperform current clinical methods by $\approx 10\%$.

To further improve upon this work, I plan to gather manual labels for 6000 sequences that do not match an always pattern and up sample sequences from notes that do not contain any keyword matches to further improve the generalizability of my model. I also want to improve the accuracy of my deep learning model to be >93%. I am in the process of implementing an active learning loop that will be used to pick particular patients and sequences by using entropy scores to label uncertain cases and UMAP clustering (McInnes, 2018) of ClinicalBERT word embeddings on the sequences of the N = 13,941 patients not included in the training, validation, or test sets. This active learning loop will be used to label the $\approx$ 45K patients in the MGH Accountable Care

Organization (ACO) system. All clinical adjudication will be performed by a team of 10 expert neurologists.

This research project is a valuable reference for future researchers in two ways:

1. The deep learning model can be fine-tuned to understand other diseases such as diabetes, hypertension, sleep apnea, cardiovascular, and other common chronic conditions as their early detection is not defined by laboratory testing but by clinical criteria found in EHR (like that of dementia). With a record 89% adoption of EHR, programming them poses a huge untapped opportunity to facilitate surveillance and management of chronic diseases.

2. The two-step NeuraHealth framework can be applied to the screening of other diseases from EHR to significantly increase accuracy of results.

In conclusion, the establishment of an automated screening pipeline to perform early detection of CI in EHR through my project provides a tool that significantly outperforms current clinical methods and can allow for the initiation of appropriate treatment to prevent complications related to dementia and save lives. Additionally, the deep learning ClinicalBERT model can be fine-tuned to understand diabetes, hypertension, sleep apnea, cardiovascular, and other pervasive chronic conditions as their early detection is not defined by laboratory testing but by clinical criteria found in EHR (like that of dementia). Therefore, it opens new avenues in the detection, management and treatment for many chronic diseases and its implementation in the healthcare system will be transformational for global health.